\documentclass[conference,compsoc]{IEEEtran}

\usepackage[ruled,linesnumbered,algo2e]{algorithm2e}
\usepackage{amsfonts}
\usepackage{amsmath,amsthm}
\usepackage{booktabs}
\usepackage[nocompress]{cite}
\usepackage[inline]{enumitem}
\usepackage{etoolbox}
\usepackage[utf8]{inputenc}
\usepackage{todonotes}
\usepackage{url}
\usepackage{xfrac}

\let\bbordermatrix\bordermatrix
\patchcmd{\bbordermatrix}{8.75}{4.75}{}{}
\patchcmd{\bbordermatrix}{\left(}{\left[}{}{}
\patchcmd{\bbordermatrix}{\right)}{\right]}{}{}

\DeclareMathOperator*{\argmin}{argmin}

\SetCommentSty{mycommfont}
\newlength\mylen

\SetKwComment{Comment}{$\triangleright$\ }{}

\begin{document}

\title{Online Soft Conformance Checking: \\ Any Perspective Can Indicate Deviations \vspace{-1em}}
\author{\IEEEauthorblockN{Andrea Burattin}
\IEEEauthorblockA{Technical University of Denmark}}
\pagestyle{plain}

\maketitle

\begin{abstract}
    Within process mining, a relevant activity is conformance checking. Such activity consists of establishing the extent to which actual executions of a process conform the expected behavior of a reference model. Current techniques focus on prescriptive models of the control-flow as references. In certain scenarios, however, a prescriptive model might not be available and, additionally, the control-flow perspective might not be ideal for this purpose. This paper tackles these two problems by suggesting a conformance approach that uses a descriptive model (i.e., a pattern of the observed behavior over a certain amount of time) which is not necessarily referring to the control-flow (e.g., it can be based on the social network of handover of work). Additionally, the entire approach can work both offline and online, thus providing feedback in real time. The approach, which is implemented in ProM, has been tested and results from 3 experiments with real world as well as synthetic data are reported.
\end{abstract}

\section{Introduction}

Process mining~\cite{VanderAalst2016} is emerging as both an important research topic in academia and a mature industrial driver~\cite{Kerremans2018}. Specifically, it aims at combining business processes and event logs in order to gather useful information and knowledge. Within process mining, three activities are typically identified~\cite{VanderAalst2016}: control-flow discovery, conformance checking, and enhancement. In this paper, we focus on conformance checking, which consists of comparing a process model, called \textit{reference model}, and an execution trace (both given as input) to establish the extent to which the observed execution conforms the given model.

In process mining, it is typical to distinguish between \textit{descriptive} and \textit{prescriptive} business process models~\cite{Carmona2018}. Descriptive models aim at showing what reality ``looks like'', i.e., they represent what actually happened in actual executions. On the contrary, prescriptive models aim at capturing what should happen, for example as imposed by regulations or protocols. In the context of conformance checking it is common to use, as reference model, a prescriptive one. Additionally, in the literature, only the control-flow perspective, i.e., how the different activities are organized, has been used for conformance checking purposes. An example of another perspective is the social one~\cite{VanDerAalst2004} which describes the relationships among resources involved in the execution of a process instance.

In this paper, we present a conformance checking technique that works on descriptive models (instead of prescriptive ones) that might refer to any perspective. In some settings, for examples in some hospital departments, it might not be particularly effective to analyze the conformance of the activities from a prescriptive model, since such reference process could be very flexible to allow physicians to decide how to behave in each and every situation. However, it might be relevant to \textit{learn} a descriptive model referring to the social perspective (i.e., how a patient with a specific disease is treated or inspected by different types of workers, such as doctors, nurses, and specialists) and then monitor for deviations of such model. We call such type of conformance, which refers to a descriptive model constructed potentially from any perspective, \textit{Soft Conformance}.
Additionally, we aim at computing the Soft Conformance in an online fashion. In the context of this paper, with ``online'' we refer to the input of the technique described: we do not assume a static event log but a \textit{stream of events}. In~\cite{Aggarwal2007,Bifet2010} a data stream is reported as an unbounded sequence of data items which are generated at very high throughput. In order to process such stream, it is necessary to meet the following important requirements:
\textit{(i)} each observation is assumed to contain just a small and fixed number of attributes;
\textit{(ii)} algorithms processing data streams need to go through a potentially infinite amount of data, without exceeding memory limits;
\textit{(iii)} algorithms have to scale linearly with the number of processed items.
Online process mining has been investigated in the past~\cite{Burattin2018}, but only to a limited extent.

The rest of the paper is structured as follow:
Section~\ref{sec:related} summarizes the state-of-the-art reporting the most relevant related work;
Section~\ref{sec:approach} presents the main approach for Online Soft Conformance;
Section~\ref{sec:demo-impl} reports the results of three different validation experiments and describes some implementation details;
finally, Section~\ref{sec:conclusion} concludes the paper.

\section{Related Work}
\label{sec:related}

Conformance checking received a lot of attention in the process mining field~\cite{Carmona2018}. Specifically, state-of-the-art approaches for conformance checking tackle the offline setting using the idea of \textit{alignments}~\cite{Adriansyah2014}, i.e., aligning a given trace with the closest trace that can be generated by the model. Due to the complexity of computing these alignments, many techniques have been devised to optimize such activity, by investigating some orthogonal directions. Such techniques include improving the search algorithm for the closest trace (based on A* search algorithm)~\cite{VanDongen2018a}, adopting planning techniques to solve the conformance problem~\cite{DeLeoni2017}, decomposing the problem into smaller fragments~\cite{Munoz-Gama2014}, and approximating the alignments~\cite{Taymouri2016}. All these approaches are not suitable for our problem because they cannot be used in online settings. Additionally, they assume, as reference model, a prescriptive one.

Online process mining~\cite{Burattin2018}, recently, started to investigate the problems related to process mining in settings where time and memory constraints play fundamental roles. Approaches for online conformance checking focused on different aspects, either by replicating the token-replay mechanism~\cite{Weber2015}, or by incrementally constructing alignments~\cite{VanZelst2017}. Few other approaches, finally, focused on the more rigorous constraints of streaming data processing, thus tackling the problem from different angles to construct measures not based on alignments~
\cite{Burattin2017a,Burattin18online}. These approaches, however, require a prescriptive model as a reference, which refers to the control-flow perspective.

Concerning checking the conformance of perspectives which are not the control-flow, only a few techniques are capable of doing that. For example, in~\cite{Mannhardt2016} authors present a technique to compute the conformance of Petri nets with data, i.e., Petri nets enriched with guards on the data perspective. Similarly, the technique in~\cite{Burattin2016a} computes the conformance of a Declare model enriched with data conditions for the constraints. Both these techniques, however, rely on the control-flow perspective as ``backbone'' of the definition of the model. These techniques cannot be used in online settings and they expect a prescriptive model as input.

Probabilistic representations of process model (as used in this paper) are the exception in the literature, and only a few papers report investigations on this regard, but not for conformance checking purposes. For example, in~\cite{Weber2011b} authors detect the ``concept drift'', i.e., when the process structure changes over time; in~\cite{Ferreira2009}, authors estimate the transition matrix from event logs without the case id.

All these approaches, in conclusion, assume a prescriptive model as reference, they all refer to the control-flow perspective (though some of these mention other perspectives), and only a few of them can be used in online settings.

\section{Online Soft Conformance}
\label{sec:approach}

The main idea of this paper is to change the typical way of seeing the conformance checking problem: instead of considering a prescriptive model, we aim at using a descriptive one. Additionally, we aim at computing the conformance in an online fashion, as opposed to the typical \textit{post-mortem} case. Conformance checking techniques that fulfill these two assumptions are called ``Online Soft Conformance'' (OSC). Such a paradigm shift on the premises of the problem imposes to re-think the entire setting. The aim of this section is to provide the used in the rest of the paper.

The first concept that needs to be clarified is what is, in our domain, a descriptive model and as explained in Section~\ref{sec:descriptive-model}. Additionally, because of its nature, a descriptive model needs to be pre-processed in order to be used for online soft conformance checking and this is covered in Section~\ref{sec:descriptive-model-conformance}. Finally, in Section~\ref{sec:soft-conformance}, we present the envisioned approach for OSC.

The goal of this paper is to devise, implement and test a technique which fulfills the following requirements:
\begin{enumerate}[label={\sffamily R\arabic{enumi}}]
    \item \label{req:perspectives} The conformance checking should not be limited to the control-flow, but other perspectives should also be investigated.
    \item \label{req:descriptive} The technique should assume a descriptive model rather than a prescriptive one.
    \item \label{req:online} The conformance checking technique should be used in offline and online settings.
\end{enumerate}

\subsection{Descriptive Model}
\label{sec:descriptive-model}

In conformance checking, the distinction between \textit{descriptive} and \textit{prescriptive models} is based on whether a model represents how the reality ``is'' or ``should be''~\cite{Carmona2018}. Leveraging such distinction, it is important to define what it means to have a model that represents how the reality is (i.e., a descriptive model). In the context of this paper, and to cope with \ref{req:perspectives} and \ref{req:descriptive}, we decided to define a descriptive model as a \textit{transition matrix} (sometimes, in the literature, also called \textit{stochastic matrix})~\cite{Privault2013,Ibe2013} identifying the probability of observing a transition from one \textit{accomplishment} to another. We decided to adopt the term ``accomplishment'' to not bind our definition to any specific perspective: an accomplishment can be the execution of a certain activity (i.e., control-flow perspective) or the processing of a work item by a given resource (i.e., organizational perspective). A transition matrix is a square matrix with one column/row for each accomplishment. The values in the matrix indicate the probability of observing the two accomplishments -- referring to the row and the column -- one after the other.
For example, let's consider the following transition matrix:
\begin{equation}
    \label{eq:ex1}
    \mathbf{T} = \bbordermatrix{
        & A & B & C \cr
        A & 0.2 & 0.8 & 0 \cr
        B & 0 & 0 & 1 \cr
        C & 0 & 0 & 1 \cr
    }
\end{equation}
It describes three ``accomplishments'' (i.e., $A$, $B$ and $C$) with corresponding probabilities, e.g., the probability of observing the execution of $B$ after $A$, denoted with $\mathbf{T}_{1,2}$ (where $\mathbf{T}$ indicates the name of the matrix, and the subscripts refer to the row/column indexes in the index set $\mathbb{S}^2 = \mathbb{S} \times \mathbb{S}$), is 0.8; wheres, using an alternative syntax, $\mathbf{T}(C, B) = 1$ (where $C$ and $B$ refer to the accomplishments).
\begin{figure}
    \centering
    \includegraphics[trim={1cm 1cm 1cm 1cm},clip,width=.4\textwidth]{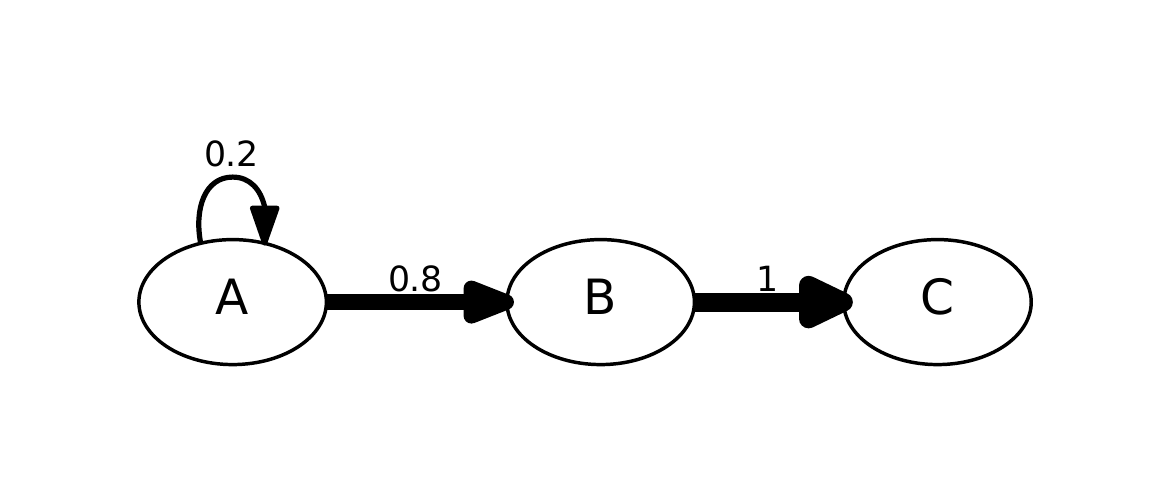}
    \caption{Graphical representation of the transition matrix reported in Eq.~\eqref{eq:ex2}.}
    \label{fig:ex1}
\end{figure}
It is important to note that a transition matrix, as defined so far, is capable of capturing only local behavior, i.e., the probability of a certain accomplishment only depends on the previous one.
Additionally, because of the probability definition, the values of each row of a transition matrix sum to 1, i.e., $\sum_{j\in\mathbb{S}} \mathbf{T}_{i,j} = 1$ for each row index $i \in \mathbb{S}$.

The assumption that the probabilities of each row sum to 1 implies that the stochastic process described by a transition matrix does not terminate. Considering the example reported in Eq.~\eqref{eq:ex1}, it is possible to note that, from accomplishment $C$, with probability 1, there will be another $C$ afterward.
This is seldom the case in the business process context: business processes are typically characterized by a goal to achieve, i.e., a termination state. Therefore, by relaxing the previously mentioned property we end up with so-called \textit{sub-stochastic matrices}~\cite{Kijima1997}. In a sub-stochastic matrix $\textbf{P}$, it is required that each probability $\mathbf{P}_{i,j} \geq 0$ and that $\sum_{j\in\mathbb{S}} \mathbf{P}_{i,j} \leq 1$ for each row index $i \in \mathbb{S}$.
An example of such matrix is the following:
\begin{equation}
    \label{eq:ex2}
    \mathbf{P} = \bbordermatrix{
        & A & B & C \cr
        A & 0.2 & 0.8 & 0 \cr
        B & 0 & 0 & 1 \cr
        C & 0 & 0 & 0 \cr
    }
\end{equation}
This matrix creates the model graphically represented in Fig.~\ref{fig:ex1}. It is possible to interpret $C$ as the final accomplishment, terminating the execution of the business process.

Please note that the structure presented here as descriptive model represents a subset of the information contained in the \textit{dependency/frequency-table} (D/F-table)~\cite{Weijters2003} which is used in several algorithms, including the Heuristics Miner.

\subsection{Obtaining a Descriptive Model}
\label{sec:obtain-descriptive}

Descriptive models can be obtained in several ways. All the process mining discovery techniques are indeed generating descriptive models, including control-flow discovery (e.g., the Heuristics Miner, the Genetic Miner, the Region-based Miner, the Inductive Miner~\cite{VanderAalst2016}) and social network mining~\cite{VanderAalst2005Social,VanDerAalst2004}. In case the model is dealing with control-flows, then it is possible to construct a descriptive model by considering the so-called \textit{directly following relationships}~\cite{Aalst2004}. In case the model is a social network then, by focusing on the ``handover of work'' metric, we obtain a descriptive model, very similar to what described in Section~\ref{sec:descriptive-model} (except for the normalization of the weights of the edges).

In the rest of this subsection, we present a possible approach to constructing a descriptive model from a set of observations, i.e., an event log~\cite{VanderAalst2016}. Given a set of attribute names $\mathcal{A}_n$ and a set of attribute values $\mathcal{A}_v$, an event $e: \mathcal{A}_n \to \mathcal{A}_v$ is a key-value relation (i.e., a partial function) mapping attribute names to corresponding values. Attributes names in $\mathcal{A}_n$, which are typically available, are \textit{name}, \textit{timestamp}, and \textit{originator}. 
An example of an event is $e = \{\textit{name = `purchase'}, \textit{timestamp = 2019-06-24}, \textit{cost = 100}\}$, with $e(\textit{name}) = `\textit{purchase}'$ and $e(\textit{cost}) = 100$. In the literature, it is not uncommon to also define a projection operator $\pi$ to extract single components of an event which, in this context can be defined as $\pi_a(e) = e(a)$ (e.g., referring to the previously defined event $e$ we would obtain $\pi_\textit{name}(e) = e(\textit{name}) = `\textit{purchase}'$). In the rest of the paper, we will use the two definitions interchangeably.
The set of all events is denoted with $E$, the set of all possible sequences of events is called $T = E^*$, and a sequence of events $t \in T$ is called \textit{trace}. Single events of $t = \langle e_1, e_2, \dots, e_n \rangle$ are accessed by the corresponding index, e.g., $t(1) = e_1, t(2) = e_2$. Additionally, let's extend the projection operator $\pi$ to traces, e.g., $\pi_\textit{name}(t) = \langle \pi_\textit{name}(e_1), \pi_\textit{name}(e_2), \dots, \pi_\textit{name}(e_n) \rangle$.

Given a set $A$, a multiset $M: A \to \mathbb{N}_0$ relates each element in $A$ with its frequency (which can be 0 or more). For example, let's consider $A = \{a_1, a_2, a_3\}$, a possible multiset is $M = \{ (a_1, 1), (a_2, 0), (a_3, 2) \}$. An alternative writing of $M$ is $M = [a_1, a_3^2]$, where the superscript reports the frequency of each element and, additionally, $a_2$ is omitted (since it has frequency 0) and for $a_1$ the frequency is omitted (since it is 1).
With these definitions in place, we call \textit{log} a multiset of traces. Additionally, let's extend the projection operator $\pi$ to logs, e.g., $\pi_\textit{name}(L) = [\pi_\textit{name}(t)]$ for all traces $t \in L$.

Considering a certain attribute \textit{a} and a log $L'$ we can extract:
$\pi_\textit{a}(L') = [\langle A,B,C\rangle ^3,  \langle A,A,B,C\rangle]$.
From this, we can construct the square matrix $\mathbf{DF}$, reporting the number of times each activity is observed directly following~\cite{Aalst2004} another one (cf. \cite[Sec. 4.1]{Weijters2003}):
\begin{equation}
\label{eq:raw-model}
    \mathbf{DF} = \bbordermatrix{
        & A & B & C \cr
        A & 1 & 4 & 0 \cr
        B & 0 & 0 & 4 \cr
        C & 0 & 0 & 0 \cr
    }
\end{equation}
Normalizing the numbers in the tables to be in the range $[0,1]$ by rows (i.e., the sum of each row should be at most 1), we obtain our descriptive model. In the example just reported, we end up with the matrix $\mathbf{P}$ in Eq.~\eqref{eq:ex2}.

\subsection{Descriptive Model for Soft Conformance}
\label{sec:descriptive-model-conformance}

The models we have considered so far are \textit{descriptive}. This means that the model is representative only of the behavior observed in the log used for its construction (called \textit{learning log}). This fact could limit the conformance in a very impactful way: if the model is not representative enough we will end with a lot of \textit{false positives}, i.e., conformance violations which are reported only due to lack of representativeness of the model. Please note that this problem does not occur if a prescriptive model is used: in this case the model is, by definition, complete and representative.

Considering the three accomplishments of the example in the previous sub-section, if no information is available to properly extract a representative descriptive model (i.e., no leaning log), then we can only extract the following \textit{stationary process}~\cite{Ibe2013} involving the three accomplishments:
\begin{equation}
\label{eq:all-obs}
    \mathbf{P'} = \bbordermatrix{
        & A & B & C \cr
        A & \sfrac{1}{3} & \sfrac{1}{3} & \sfrac{1}{3} \cr
        B & \sfrac{1}{3} & \sfrac{1}{3} & \sfrac{1}{3} \cr
        C & \sfrac{1}{3} & \sfrac{1}{3} & \sfrac{1}{3} \cr
    }
\end{equation}
In this case, the probability of an accomplishment does not depend on the previous one: all of them share the same probability distribution over time. To make an analogy with process mining, we could say this is a ``flower model''~\cite{VanderAalst2016} where everything has the same likelihood to occur. Therefore, the model $\mathbf{P'}$ is the most general one, in the sense that it is the least affected by observations, though it equally supports all possible direct following relations.

To tackle the representativeness problem of descriptive models we need to support behavior never observed in the learning log. The idea is to ``merge'' the model extracted from the learning log with the model which equally supports all observations. While merging the two models, we want to control which one is more important and for that we use a weighting factor $0 \leq \alpha \leq 1$, which allows us to construct the stochastic matrix $\mathbf{S}$:
\begin{equation}
\label{eq:weight}
    \mathbf{S} = \alpha\mathbf{P} + (1-\alpha)\mathbf{P'}
\end{equation}
Please note that, in case $\alpha = 0$ than $\mathbf{S}$ is exactly equal to $\mathbf{P'}$, i.e., only the flower model is considered and the learning log is ignored; in case $\alpha = 1$ than $\mathbf{S}$ is exactly equal to $\mathbf{P}$, i.e., only the descriptive model extracted from the learning log is considered and the behavior of the flower model is completely ignored. The rationale behind this idea is to have control over the amount of additional behavior (not contained in the learning log) to support. For example, if we know that the learning log is very representative, then we can decide to ignore any relation not explicitly supported, by setting $\alpha = 1$. Instead, if we have the impression that the descriptive model is not representative enough, then we can decide on a slightly lower value of $\alpha$ thus admitting, to some extent, unseen behavior.

As example, let's consider the $\mathbf{DF}$ model in Eq.~\eqref{eq:raw-model} which is normalized into $\mathbf{P}$ as in Eq.~\eqref{eq:ex2}. Let's also assume to use a weighting factor $\alpha = 0.5$, we then obtain:
$$
    \mathbf{S} = 0.5 \begin{bmatrix}
        0.2 & 0.8 & 0 \cr
        0 & 0 & 1 \cr
        0 & 0 & 0 \cr
    \end{bmatrix} + (1 - 0.5) \begin{bmatrix}
        \sfrac{1}{3} & \sfrac{1}{3} & \sfrac{1}{3} \cr
        \sfrac{1}{3} & \sfrac{1}{3} & \sfrac{1}{3} \cr
        \sfrac{1}{3} & \sfrac{1}{3} & \sfrac{1}{3} \cr
    \end{bmatrix}
$$\begin{equation}
\label{eq:final}
    \mathbf{S} = \bbordermatrix{
        & A & B & C \cr
        A & 0.26 & 0.57 & 0.17 \cr
        B & 0.17 & 0.17 & 0.66 \cr
        C & 0.17 & 0.17 & 0.17 \cr
    }
\end{equation}
\begin{figure}
    \centering
    \includegraphics[trim={1cm 1cm 1cm 1cm},clip,width=.4\textwidth]{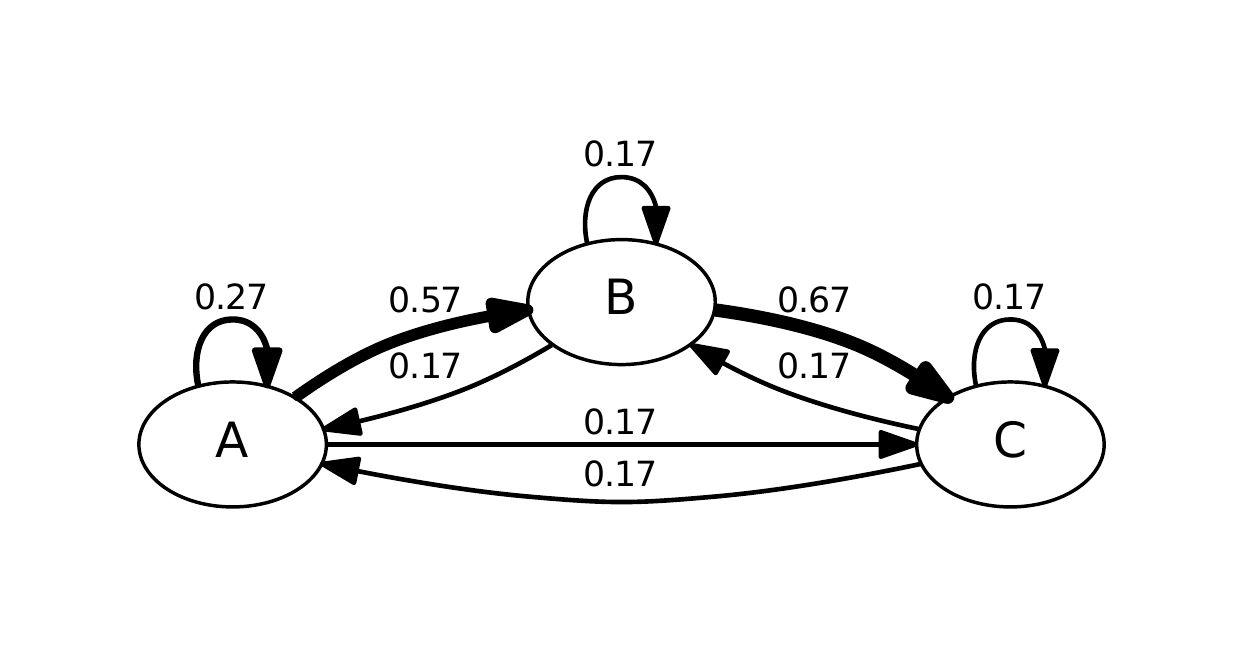}
    \caption{Graphical representation of the transition matrix of the running example (from Eq.~\eqref{eq:ex2}) after being merged with the model supporting all observations (from Eq.~\eqref{eq:all-obs}) with weighting factor $\alpha = 0.5$, as in Eq.~\eqref{eq:final}.}
    \label{fig:ex2}
\end{figure}
A graphical representation of this model is reported in Fig.~\ref{fig:ex2} (numbers are rounded to 2 decimals). Please note that since the descriptive model (i.e., Eq.~\ref{eq:ex2}) is sub-stochastic, $\mathbf{S}$ is sub-stochastic as well.

% In the rest of the paper, for readability purposes, we assume to access the values of matrix $\mathbf{S}$ using the following syntax $\mathbf{S}(a_i,a_j)$ where $a_i$ and $a_j$ refers to accomplishments of the matrix. For example, considering the matrix in \eqref{eq:final}, $\mathbf{S}(A,B)=0.57$ and $\mathbf{S}(B,C)=0.66$.

\subsection{Online Soft Conformance Checking}
\label{sec:soft-conformance}

This section presents the approach for computing the online soft conformance checking, once a proper descriptive model has been prepared.

First of all, given the first $n$ positive natural numbers $\mathbb{N}_n^+ = \{ 1, 2, \dots, n \}$ and a target set $A$, a \textit{sequence} $\sigma$ is a function $\sigma: \mathbb{N}_n^+ \to A$. We say that $\sigma$ maps indexes to corresponding elements of the target set $A$, for example if $\sigma = \langle a_1, a_2, \dots, a_n\rangle$, then $a_i = \sigma(i)$ with $a_i \in A$.
Then, given a set of accomplishments $\mathcal{A}$ (e.g., the tasks of a process model, the originators involved in activities), a \textit{trace} $T$ of length $n$ is a sequence $T : \mathbb{N}_n^+ \to \mathcal{A}$. Accomplishments that belong to the same process instance are grouped into the same trace.
Finally,  given the \textit{event universe} $\mathcal{E} = \mathcal{A} \times \mathcal{C}$, where $\mathcal{A}$
is the set of accomplishments and $\mathcal{C}$ is the set of possible case ids, an event stream $\Psi$ is
an infinite sequence $\Psi : \mathbb{N}^+ \to \mathcal{E}$. Please note that in an event stream contiguous events may belong to different process instances, i.e., all traces are intertwined each other.

With an event stream available it is possible to extract the stream of directly following relations belonging to the same process instance (for example, as reported in~\cite{Burattin2017a}) and it is possible to extract corresponding probabilities from a descriptive model properly prepared (as described in Sec.~\ref{sec:descriptive-model-conformance}). The sequence of probabilities obtained can be grouped in different ways: for example, it is possible to multiply probabilities within the same process instance. This approach, however, penalizes executions just based on their lengths rather than their conformance levels. For this reason, we decided to combine the sequence of probabilities into their mean (per trace). Additionally, we normalized the mean in order to compensate the sub-stochasticness introduced by the weight factor $\alpha$ as explained in Sec.~\ref{sec:descriptive-model-conformance}. Such normalized mean realizes the Soft Conformance measure.

\begin{algorithm2e}[t]
	\caption{Online Soft Conformance Checking \label{alg:conformance}}
	\scriptsize
	\DontPrintSemicolon
	\KwIn{$\mathbf{S}$: a prepared descriptive model (as described in Sec.~\ref{sec:descriptive-model-conformance}) \newline
        $\alpha$: the weighting factor used to prepare $\mathbf{S}$ (cf. Sec.~\ref{sec:descriptive-model-conformance}) \newline
	    $\mathcal{A}$: the set of accomplishments in $\mathbf{S}$ \newline
		$\Psi$: event stream \newline
		$m$: maximum number of parallel instances}
	\DontPrintSemicolon
	\BlankLine
	$M : \mathcal{C} \to \mathcal{A} \times \mathbb{R}^+ \times \mathbb{N}^+ \times \mathbb{N}^+$ \label{alg:conf:map} \tcp*{Hash map which, given a case id, returns a tuple with the latest accomplishment observed, the current mean of the probabilities, the number of accomplishments observed on the given trace, and the time of last update}
	\For{\hspace{-.3em}\emph{\textbf{ever}} \label{alg:conf:loop}}{
		$(a, c) \gets \textit{observe}(\Psi)$ \label{alg:conf:observe} \tcp{New event containing the accomplishment and the case id}
		\BlankLine
		\tcp{Update the data structures}
		$(\textit{acc}, \textit{mean}, \textit{obs}, \textit{time}) \gets M(c)$ \;
		\uIf(\tcp*[h]{First accomplishment of trace (so no transition yet)}){$(\textit{acc}, \textit{mean}, \textit{obs}, \textit{time}) = \bot$ \label{alg:conf:if1}}{
			$M(c) \gets (a, 0, 0, \textit{now})$ \label{alg:conf:noobs} \;
		}
		\Else (\tcp*[h]{We have a transition, we need to update the mean of the probabilities}) {
		    $\textit{mean} \gets \textit{mean} + \dfrac{\mathbf{S}(\textit{acc}, a) - \textit{mean}}{\textit{obs} + 1}$ \label{alg:conf:newmean1} \;
		    $M(c) \gets (a, \textit{mean}, \textit{obs} + 1, \textit{now})$ \label{alg:conf:newmean2} \;
		}
		\BlankLine
		Notify new Soft Conformance for $c$: $\dfrac{\textit{mean}}{\alpha + \dfrac{1 - \alpha}{|\mathcal{A}|}}$ \label{alg:conf:sc} \;
		\tcp{Cleanup the map}
		\If{$|M| > m$}{
			$R \gets \argmin_{(a, m, o, t) \in M} \{ t \} $ \label{alg:conf:selectold} \tcp{Get oldest elements in $M$}
			Remove $R$ from $M$ \label{alg:conf:dropold} \;
		}
		\label{alg:conf:end}
	}
\end{algorithm2e}
The pseudo-code for computing the Online Soft Conformance is reported in Algorithm~\ref{alg:conformance}.
It starts by constructing a hash map that will keep track of the most recent process instances (line~\ref{alg:conf:map}).
Then, the algorithm begins the actual online procedure by repeating forever (line~\ref{alg:conf:loop}) the main loop which starts with the observation of a new event from the stream (line~\ref{alg:conf:observe}). This event refers to one accomplishment and a case id.
After that, the mean of the probabilities of transitions needs to be updated. Specifically, if the accomplishment is the first of the trace (lines~\ref{alg:conf:if1}-\ref{alg:conf:noobs}), then no transitions were observed. Otherwise (i.e., for the current case id other accomplishments were observed before), the algorithm updates and stores the new mean (lines~\ref{alg:conf:newmean1}-\ref{alg:conf:newmean2}).
Once these updates are complete, it is possible to notify the new value of the Soft Conformance (line~\ref{alg:conf:sc}). This value is calculated as the mean of the transitions, normalized in the range $[0,1]$. To perform such normalization we divide the mean by the maximum value that can be obtained (i.e., when the descriptive model contained a probability 1) after the weighting reported in Eq.~\eqref{eq:weight}. This value is actually $\alpha \cdot 1 + (1 - \alpha) \cdot \frac{1}{\text{no. of accomplishments}}$ which can be simplified as reported in line~\ref{alg:conf:sc} of the algorithm.

\vspace{1em}
At the beginning of Sec.~\ref{sec:approach}, 3 requirements were presented for the devised technique. \ref{req:perspectives} refers to the possibility of using any perspective for the conformance. This is indeed the case as the reference model can be constructed starting from any general accomplishment. 
\ref{req:descriptive} mentioned that OSC should use a descriptive model instead of a prescriptive one. As reported in Sec.~\ref{sec:descriptive-model-conformance}, after pre-processing a descriptive model, it is indeed used for the actual conformance.
\ref{req:online} requires the suitability of the approach for offline and online computation. With the online case being the most restrictive one, the algorithm has to show constant time complexity for each observation processed (i.e., within lines~\ref{alg:conf:observe}-\ref{alg:conf:end})~\cite{Gama2010}. This is indeed the case: the only data structure used is a hash map which requires constant time for corresponding operations. The other steps involve just simple arithmetic calculations (which require constant time complexity) and retrieving the probability of a transition from $\mathbf{S}$ (in line~\ref{alg:conf:newmean1}). Since the transition matrix has constant size (it does not change w.r.t. the input), fetching such probabilities can also be completed in constant time.
For these reasons, the theoretical computational complexity of the algorithm (which is constant per event processed) makes it a viable solution for online applications (and, as a consequence, for offline as well) and therefore \ref{req:online} is also fulfilled.

\section{Demonstration and Implementation}
\label{sec:demo-impl}

This section presents the results of 3 tests where we successfully applied the Soft Conformance as well as some implementation details. The first test reports an original scenario, referring to real data, where it would be impossible to gain insights using standard conformance checking techniques. The second test compares the results of the Soft Conformance and standard state-of-the-art techniques, showing that these two correlate, indicating the stability of the measure. The third test aims at verifying the robustness of the prototype under stress conditions.

\subsection{Use Case Scenario: Eye Tracking}

A relevant real-world application of the technique presented in this paper is in the context of eye tracking~\cite{Holmqvist2011}. Specifically, with eye tracking it is possible to virtually divide a screen into Areas Of Interest (AOIs) and track, over time, where a subject is focusing their attention in response to a stimulus (e.g., a question to answer). It is possible to use \textit{fixations} on AOIs (i.e., a certain amount of time spent looking at an AOI) as an approximation of cognitive activities~\cite{Andaloussi2018,Weber2016}. For example, if a subject is focusing on a text, we can assume that the person is reading it.

In~\cite{Andaloussi2018}, authors leveraged eye tracking and process mining to analyze how subjects were solving a sense-making exercise where different types of aid were available (i.e., a graph, a simulation tool, and a text). Authors were also able to extract three different ``profiles'' for subjects, based on the different interactions among the AOIs (i.e., ``graph profile'', ``simulation profile'', ``text profile'').
For this paper we would like to use the same context and the same data to push the investigation a bit further: we will use a profile as a descriptive model and assess the conformance of all other traces w.r.t. it. In this case, it is not possible to use a prescriptive model: there is no ``correct'' or ``incorrect'' way of solving the exercise and, even within the same ``profile'', the executions might contain differences.

First, we need to construct our descriptive model, which captures a certain profile. To do that, we attributed the subjects to graph, simulation, or text profiles considering the relative amount of time spent on inspecting the three AOIs. Percentages are reported in Table~\ref{tbl:class}. With these percentages, it is possible to classify subjects based on the AOI that received most of their attention. In~\cite{Andaloussi2018}, the different maps for each profile are reported, showing different interactions with the AOIs.
\begin{table}
    \centering
    \caption{Subjects used for the eye tracking experiment, with the corresponding distribution of time over different AOIs and Soft Conformance measures for graph profile.}
    \label{tbl:class}
    \begin{tabular}{l|rrr|r}
        \toprule
        & \multicolumn{3}{c|}{\textbf{Time distribution per AOI}} \\
        \textbf{Subj.} & \textbf{Graph} & \textbf{Text} & \textbf{Simulation} & \textbf{Soft Conformance} \\
        \midrule
        S01 & 20\% & \textbf{80\%} & 0\% & \textcolor{gray}{0.51} \\
        S02 & 47\% & 1\% & \textbf{52\%} & \textcolor{gray}{0.50} \\
        S03 & 44\% & 0\% & \textbf{56\%} & \textcolor{gray}{0.48} \\
        S04 & \textbf{100\%} & 0\% & 0\% & \textbf{0.72} \\
        S05 & 37\% & \textbf{42\%} & 21\% & \textcolor{gray}{0.48} \\
        S06 & 29\% & 1\% & \textbf{70\%} & \textcolor{gray}{0.47} \\
        S07 & \textbf{100\%} & 0\% & 0\% & \textbf{0.72} \\
        S08 & \textbf{52\%} & 20\% & 28\% & Used for learning \\
        S09 & \textbf{94\%} & 3\% & 3\% & Used for learning \\
        \bottomrule
    \end{tabular}
\end{table}

For our experiment, we decided to focus on the graph profile users, i.e., S04, S07, S08, S09. Additionally, we split the users and we considered S08 and S09 as ``learning log'', i.e., used them to construct the descriptive model. We merged such descriptive model with the most general one by using a weighting factor $\alpha = 0.99$. With the resulting model, we calculated the Soft Conformance for all the other subjects. Values are reported in Table~\ref{tbl:class}. It is interesting to note that subjects S04 and S07 obtained the highest Soft Conformance score which is in line with our expectations: these two are also supposed to belong to the graph profile, like those used for learning.

Despite the very low number of subjects and observations, this test proves the feasibility and the meaningfulness of Soft Conformance in real and original contexts. Also, in such scenario, since the models of the three profiles are topologically very similar between each other~\cite{Andaloussi2018}, it would make no sense to apply standard conformance checking techniques: each subject might achieve their goal in a different way, which shows only some commonalities with the reference model. For this reason, a prescriptive model would not suitable in this case.

This experiment shows that the presented technique can be used on accomplishments that are not directly coming from the control-flow perspective but, as presented in this case, are coming from an eye tracker, thus proving~\ref{req:perspectives}. Additionally, since the model used is a pure descriptive model, we could also test the application of~\ref{req:descriptive}.

\subsection{Correlation with Offline Approaches}

For the second experiment, we investigated the stability of the Soft Conformance with respect to the standard offline alignment-based conformance checking~\cite{Adriansyah2014}. We considered a real event log publicly available, called ``NASA Crew Exploration Vehicle (CEV) Software Event Log''\footnote{\label{foot:nasa}See \url{https://doi.org/10.4121/uuid:60383406-ffcd-441f-aa5e-4ec763426b76}.}. This log contains an ``\textit{event log contains method-call level events describing a single run of an exhaustive unit test suite for the Crew Exploration Vehicle (CEV) example available and documented}''\textsuperscript{\ref{foot:nasa}}.

Since the technique presented in this paper requires the construction of a descriptive model, we decided to consider the activities as ``accomplishments'' and then we divided the log into a learning log and a validation log. The learning log contains 66 process instances summing up to 904 events. The rest of the log, which has been used for validation, contains 2500 process instances for a total of 35915 events. For selecting the learning traces, we used the tool Disco\footnote{See \url{http://fluxicon.com/disco/}.} and we filtered the log to preserve only the most frequent variants.
%
% \begin{figure*}
%     \centering
%     \includegraphics[angle=90,width=\textwidth]{figures/model.pdf}
%     \caption{Petri net extracted with Inductive Miner from the learning log of the NASA dataset. For readability purposes, the picture is rotated by 90 degrees.}
%     \label{fig:nasa-pn}
% \end{figure*}
%
With the two logs available, we mined the model with the Inductive Miner (with its default configuration of the parameters)~\cite{Leemans2014} using the learning model. %, thus obtaining the Petri net depicted in Fig.~\ref{fig:nasa-pn}.
This model has been used to compute the offline conformance on the validation log.

We constructed a descriptive model using the technique presented in Sec.~\ref{sec:obtain-descriptive} and then we constructed several conformance models using different weight factors, i.e., giving different importance to the observations and to the incompleteness assumption ($\alpha$ from 0 to 1 with steps of 0.25; cf. Sec.~\ref{sec:descriptive-model-conformance}). For each of these models we compared the final value (i.e., the value at the end of the trace) of the Soft Conformance with the trace fitness and the raw fitness costs computed with standard offline techniques based on alignments~\cite{Adriansyah2014} and using the reference model extracted with the Inductive Miner from the learning log.
\begin{table}
    \centering
    \caption{Correlation between the Soft Conformance (with different weighting factors) and the Offline Trace Fitness and the Soft Conformance and the Offline Raw Fitness Cost computed using standard alignments techniques. All correlations are statistically significant.}
    \label{tab:correlation}
    \begin{tabular}{l|cc|cc}
        \toprule
        & \multicolumn{2}{c|}{\textbf{Trace Fitness}} & \multicolumn{2}{c}{\textbf{Raw Fitness Cost}} \\
        \textbf{Weight factor} & \textbf{Pearson's} $r$ & $p$\textbf{-value} & \textbf{Pearson's} $r$ & $p$\textbf{-value} \\
        \midrule
        $\alpha=0$ & 0.581 & $<.001$ & -0.355 & $<.001$ \\
        $\alpha=0.25$ & 0.687 & $<.001$ & -0.428 & $<.001$ \\
        $\alpha=0.5$ & 0.700 & $<.001$ & -0.438 & $<.001$ \\
        $\alpha=0.75$ & 0.705 & $<.001$ & -0.441 & $<.001$ \\
        $\alpha=1$ & 0.708 & $<.001$ & -0.443 & $<.001$ \\
        \bottomrule
    \end{tabular}
\end{table}
By computing the correlation between the Soft Conformance and the trace fitness and between the Soft Conformance and the raw fitness cost, we obtained the values reported in Table~\ref{tab:correlation}. First, it is interesting to observe that all correlations are statistically significant (i.e., $p < 0.001$). Secondly, both measures correlate with the Soft Conformance; the trace fitness in a positive way, the raw fitness cost in a negative way. This aligns with expectations: Soft Conformance and trace fitness values of 1 indicate no deviation; whereas the raw fitness cost increases with the number of deviations. Thirdly, the correlation with trace fitness (resp., raw fitness cost) increases (resp., decreases) as the value $\alpha$ increases. This is also in line with expectations: since the trace fitness and the raw fitness cost are computed against the reference model obtained from the learning log, the descriptive model (which also comes from the learning log) is reliable and representative. This, in turn, means that the more weight is assigned to the descriptive model, the more the two offline values resemble the Soft Conformance.

This experiment shows the feasibility of using the presented technique also in offline settings, as required by~\ref{req:online}.

\subsection{Stress test}

\begin{figure*}
    \centering
    \includegraphics[width=\textwidth]{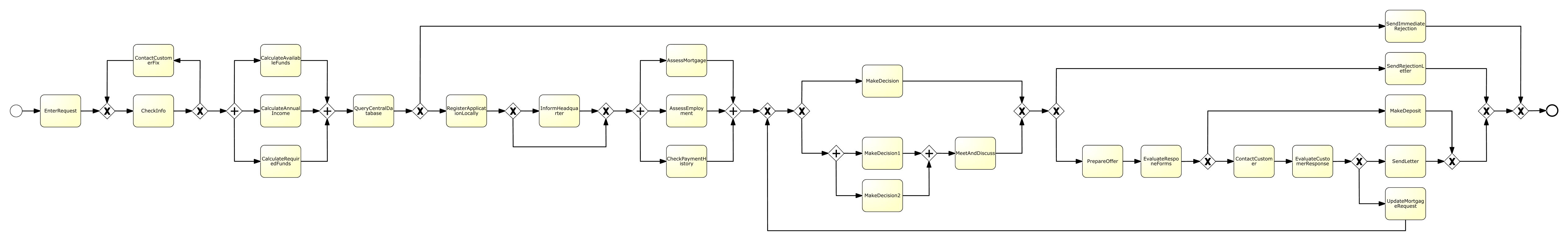}
    \caption{Mortgage process model used for simulation of the stress-test.}
    \label{fig:stress-model}
\end{figure*}

The final experiment presented in this paper aimed at proving the possibility of using the technique in online settings, as required by~\ref{req:online}. To this end, we simulated the realistic mortgage process model depicted in Fig.~\ref{fig:stress-model} with the tool PLG2~\cite{Burattin}.
The test was performed on a standard laptop, equipped with Java 1.8(TM) SE Runtime Environment on Windows 10 64bit, an Intel Core i7-7500U 2.70GHz CPU and 16GB of RAM.
\begin{figure}
    \centering
    \includegraphics[width=.5\textwidth]{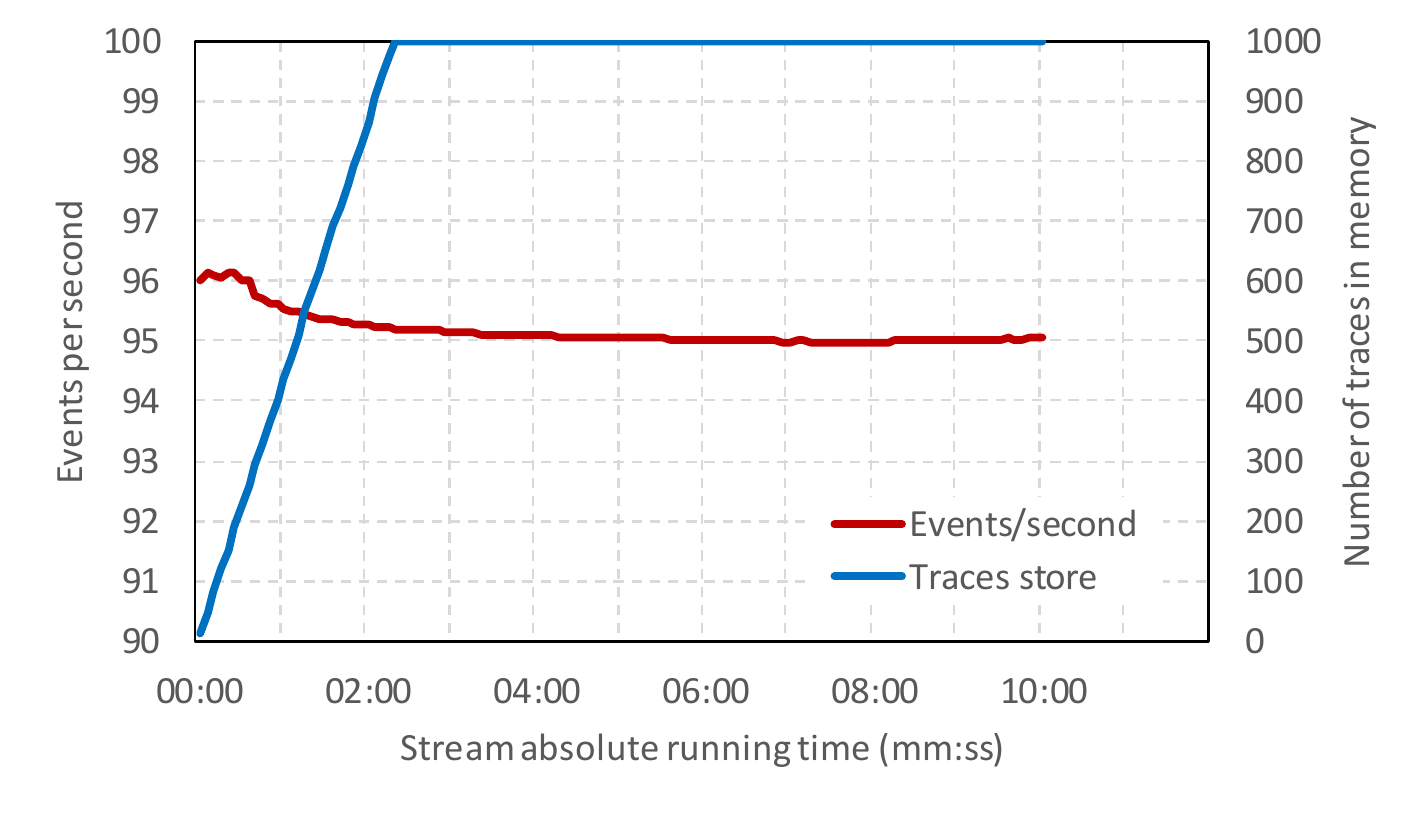}
    \caption{Results of the stress test.}
    \label{fig:stress}
\end{figure}
We set the processing to support up to 1000 traces (cf. parameter $m$ of Alg.~\ref{alg:conformance}). After 10 minutes of simulation, we obtained the results plotted in Fig.~\ref{fig:stress}. As reported in the chart, the system was capable of durably processing about 95 events per second. Additionally, considering the implementation is an un-optimized prototype and the setting adopted was not tailored to the specific test, we conclude that the requirement~\ref{req:online} is met also on the prototype level.

\subsection{Implementation}

A prototype realizing the technique described in this paper has been implemented as a ProM~\cite{Verbeek2010} package and its source code is publicly available\footnote{See \url{https://github.com/delas/OnlineSoftConformance}.}.

The package implements 5 plugins. The first plugin enables the construction of a descriptive model from an event log using any accomplishments available. Another plugin converts a Handover of Work social network~\cite{VanderAalst2005Social} into a descriptive model. The third plugin enables to normalize a descriptive model using the technique reported in Sec.~\ref{sec:descriptive-model-conformance}.
% \begin{figure}
%     \centering
%     \includegraphics[width=.5\textwidth]{figures/screenshot.png}
%     \caption{Screensot of the dashboard of the Online Soft Conformance Checking ProM plugin.}
%     \label{fig:screenshot}
% \end{figure}
The two remaining plugins enable the actual calculation of the Soft Conformance. One plugin consumes a descriptive model and an event stream (provided as a TCP/IP connection) and in real time computes the Soft Conformance for all events, resulting in a dashboard similar to those provided for other online conformance checking techniques~\cite{Burattin2017b}. %A screenshot of such dashboard is depicted in Fig.~\ref{fig:screenshot}.
\begin{figure}
    \centering
    \includegraphics[width=.5\textwidth]{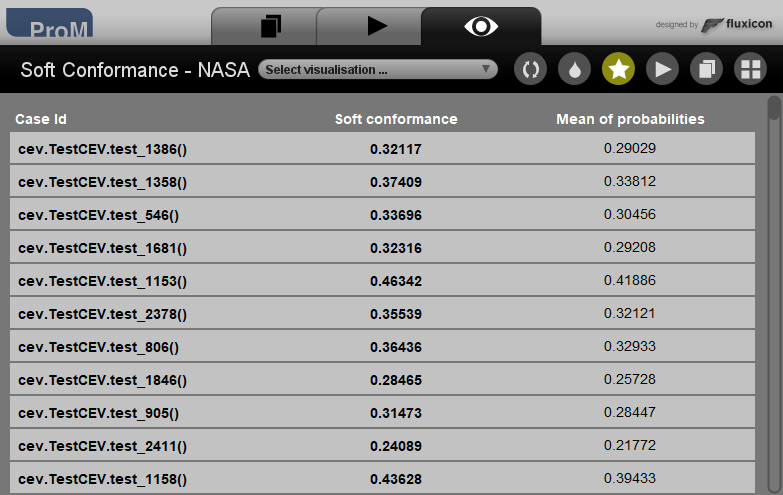}
    \caption{Screensot of the ProM plugin to compute the Soft conformance in offline fashion.}
    \label{fig:screenshot2}
\end{figure}
The final plugin calculates the Soft Conformance for an (offline) event log, cf. screenshot in Fig.~\ref{fig:screenshot2}, and reports the final value for each case id.

\section{Conclusion and Future Work}
\label{sec:conclusion}

This paper presented the Online Soft Conformance. This is the first conformance checking technique explicitly devised to use a descriptive model, instead of a prescriptive, as a reference model. Additionally, since just a descriptive model is needed, the conformance can be computed using any perspective, not just the control flow (for example, it is possible to compute the conformance w.r.t. the Handover of Work Social Network). Finally, both online and offline settings can be tackled. These three aspects are captured by corresponding requirements presented in the paper (i.e., \ref{req:perspectives}, \ref{req:descriptive}, and~\ref{req:online}) which are also empirically exhibited in the 3 experiments reported.

Despite the work presented in this paper already reports a comprehensive technique, there are several possibilities for improving it. First of all, it is important to notice that the way the descriptive model is represented allows it to capture only local behavior (i.e., directly following relationships), improving this aspect could be extremely important. Additionally, because of the definition of descriptive model, no duplicated accomplishments are allowed (i.e., just one node for each of them), though it might be very interesting and useful to cope with this limitation as well.

\subsubsection*{Acknowledgments}

This Work is supported by the Innovation Fund Denmark project EcoKnow (7050-00034A).

\bibliographystyle{IEEEtran}
\bibliography{references.bib}

\end{document}